\documentclass[11pt]{article}

\usepackage[final]{acl}

\usepackage{times}
\usepackage{latexsym}
\usepackage{makecell}
\usepackage{array}
\usepackage{booktabs}

\usepackage[T1]{fontenc}

\usepackage[utf8]{inputenc}

\usepackage{microtype}

\usepackage{inconsolata}

\usepackage{graphicx}

\usepackage{amsmath}
\usepackage{algorithm}
\usepackage{algorithmicx}

\usepackage{algpseudocode}
\usepackage{amssymb}
\usepackage{mathtools}
\usepackage{amsthm}
\usepackage{multirow}
\usepackage{placeins}

\title{Toollery: Scaling LLM Agents to Thousands of Skills and Tools}

\author{Xiangxi Tian \and Ran Guan \\
  2012 Laboratories, Huawei \\
  \texttt{\{xiangxitian1,guanran\}@huawei.com}
  }

\begin{document}
\maketitle
\begin{abstract}

As LLM agents are exposed to hundreds to tens of thousands of skills, tools, and API functions, full-library prompting becomes costly, slow, and less reliable: each added candidate increases prompt tokens and latency, while longer candidate lists introduce more distractors for LLM selection. We present \textbf{Toollery}, a training-free candidate-compression framework for scalable LLM skill/tool selection. Following established document-side query expansion, Toollery generates user-intent queries from each skill/tool specification and builds a retrieval index that maps real user requests to compact candidate sets before final LLM decision-making. By treating high-level skills and atomic tools as selectable capabilities, Toollery can be applied to both skill libraries and tool registries. We evaluate Toollery on the roughly 79K-capability SkillRouter benchmark, BFCL-V4 with over 440 atomic tools, and 3,396 proprietary smart-cockpit requests over 220 tools. Across these settings, Toollery keeps online selection bounded to a compact top-$k$ candidate set and improves recall over ordinary specification retrieval. At a fixed top-10 budget, Toollery improves end-to-end selection on the cockpit dataset, and maintains comparable AST Accuracy on BFCL-V4. These results support Toollery as a practical candidate-compression framework for large and evolving agent capability libraries, while showing that quality and cost gains depend on workload coverage and provider caching.

\end{abstract}

\begin{figure}[t]
    \centering
    \includegraphics[width=0.9\linewidth]{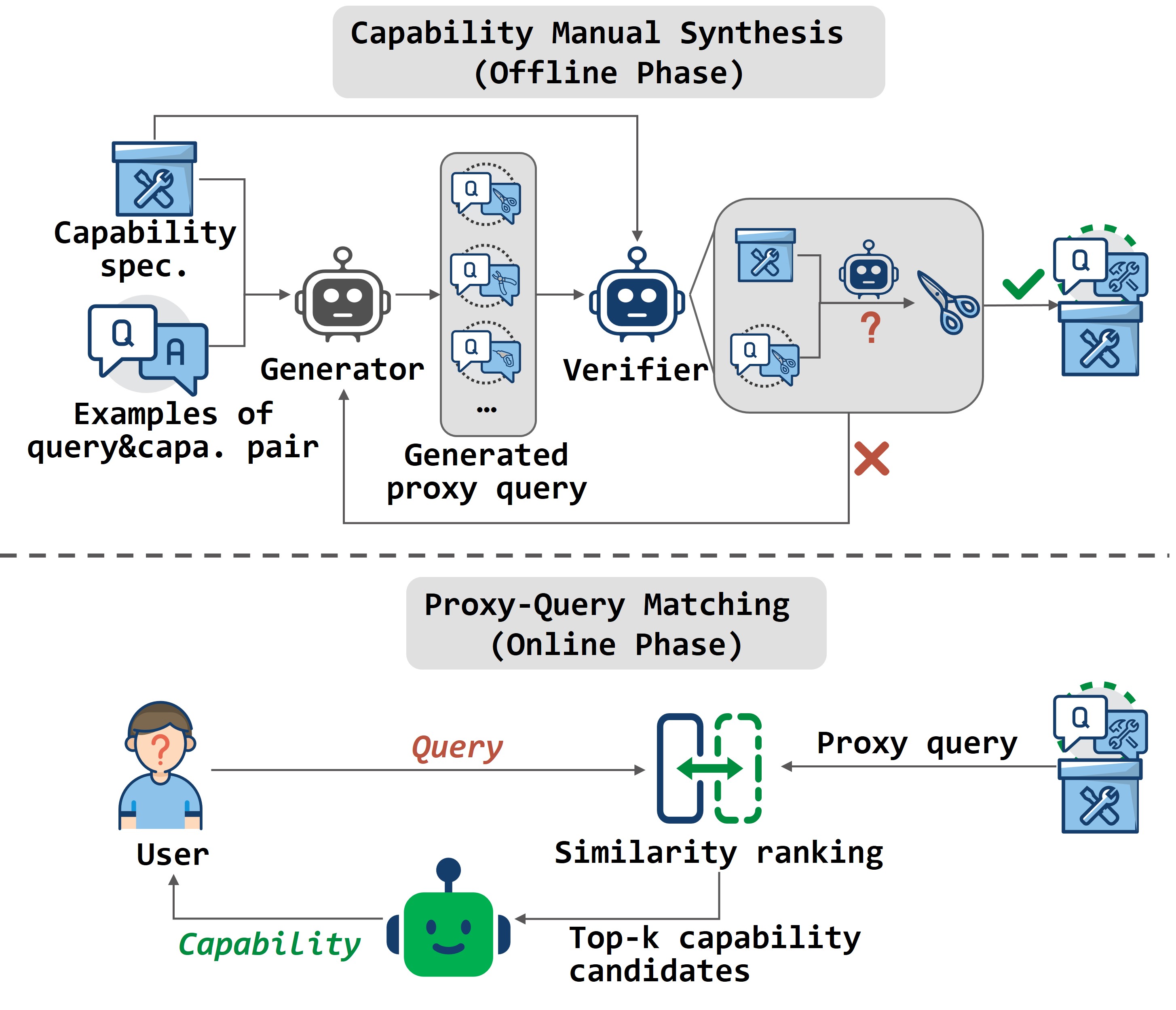}
    \vspace{-0.5em}
    \caption{\textbf{Overview of Toollery.} Toollery constructs a manual by generating user-intent queries from each capability specification. At inference time, it retrieves a compact candidate set using the manual and passes only the retrieved candidate specifications to the final LLM for decision making.}
    \label{fig:toollery-overview}
    \vspace{-0.5em}
\end{figure}

\section{Introduction}
\label{sec:intro}

LLM agents increasingly rely on external resources to complete user requests: a resource may be a high-level procedural skill, an MCP service, an atomic tool, or an API function \citep{schick2023toolformer, qin2023toolllm, patil2024gorilla}. Although these resources differ in granularity and execution interface, they play the same role before execution: the agent must decide whether each one should be exposed to the final LLM for the current user request. We therefore treat them as selectable \emph{capabilities} throughout this paper. As capability libraries grow, skill/tool selection becomes a production bottleneck. Industrial reports and our own agent deployments show that loading skill/tool specifications upfront can consume tens of thousands of tokens before any user work begins, while selection accuracy degrades as the library grows \citep{anthropic_tool_search, anthropic_advanced_tool_use}. In our deployment, after loading more than 200 skills and tools, even a trivial greeting can consume more than 50K input tokens and over 5 seconds of response latency. Scalable LLM skill/tool selection is therefore necessary for controlling token cost, latency, and accuracy as public ecosystems aggregate increasingly large capability libraries \citep{clawhub, lin2025masstool, mo2025livemcpbench, xu2025tps}.

The common practice is full-library prompting: place every skill/tool specification in the prompt and let the LLM choose. This preserves the full decision context, but makes the agent pay the cost of reviewing every available capability on every request, even when the request needs no tool. As a result, easy or irrelevant requests still incur high prompt cost, higher latency, and more distractors for final LLM decision making. Practical agents therefore need a candidate-compression layer that narrows the action space before LLM inference.

Retrieval is a natural way to implement candidate compression, but ordinary specification retrieval is limited. Existing retrieval-based methods commonly retrieve candidates directly from skill/tool specifications, including names, descriptions, and schemas \citep{lin2025masstool, zheng2026skillrouter}. This assumes that user requests resemble specifications. In practice, users describe goals, constraints, or partial tasks, while specifications describe capabilities, parameters, and developer-facing usage details. This intent--specification mismatch is especially problematic for heterogeneous libraries whose capabilities are written at different levels of abstraction.

We present \textbf{Toollery}\footnote{Code: \url{https://github.com/XiangxiTian/toollery}}, a training-free framework that frames scalable LLM skill/tool selection as \textbf{intent-level retrieval}. Rather than retrieving over specifications directly, Toollery asks what kinds of user requests should select each capability. It treats high-level skills, MCP services, atomic tools, and API functions as selectable capabilities under the same abstraction. Toollery generates user-intent queries from each capability specification, retrieves a compact top-$k$ candidate set using these intent-augmented representations, and forwards only the selected candidates to the final LLM. This separates library-scale matching from schema-aware final LLM decision making.

Our experiments show that Toollery turns the growing skill/tool-selection bottleneck into a bounded retrieval problem. On the SkillRouter benchmark, BFCL-V4, and a proprietary smart cockpit dataset, candidate pools range from hundreds of tools to roughly 79K capabilities, yet Toollery keeps online LLM selection bounded to a compact top-$k$ candidate set. This reduces the number of specifications exposed to the final LLM, keeping token cost and latency controllable as the library grows. At the same time, the retrieved top-$k$ sets achieve higher recall than ordinary specification retrieval; under the same top-10 candidate budget, Toollery improves final tool-set selection on the cockpit dataset while maintaining comparable BFCL AST Accuracy.

This paper makes the following contributions:
\begin{itemize}
    \item We treat high-level skills, MCP services, atomic tools, and API functions as selectable capabilities under a unified abstraction, and study their selection through intent-level candidate compression.
    \item We present Toollery, a training-free framework that generates user-intent queries offline and retrieves compact top-$k$ candidate sets for final LLM decision making.
    \item We evaluate Toollery across large-scale skill routing, public atomic-tool benchmarks, and proprietary vehicle-control data, showing higher retrieval recall, reduced prompt context, and improved fixed-budget cockpit selection.

\end{itemize}

\section{Related Work}

Prior work improves LLM tool use by training or evaluating models that invoke external tools once relevant tools are available in context or training data \citep{yao2022webshop, schick2023toolformer, patil2024gorilla, qin2023toolllm, li-etal-2023-api, patilberkeley}. Planning-oriented systems such as PLUTO and ToolGen address related orchestration problems \citep{huang-etal-2024-planning,wang2025toolgen}; Toollery instead addresses selection of a compact candidate set before final LLM decision-making.

Recent retrieval and routing methods reduce the action space before tool selection \citep{lin2025masstool, zheng2026skillrouter, paramanayakam2025less}. Re-Invoke and related query-generation work generate tool-use examples for retrieval \citep{chen-etal-2024-invoke,kachuee-etal-2025-improving}; MassTool and SkillRouter learn or combine retrieval signals at scale; and RAG-MCP-style systems retrieve tool specifications \citep{gan2025rag}. Dynamic ReAct and dynamic system instructions expose tools selectively during reasoning \citep{gaurav2025dynamic,franko2026dynamic}. MCP Compressor instead controls runtime schema disclosure and is complementary to candidate retrieval; we do not claim superiority without a matched end-to-end comparison \citep{atlassian2026mcpcompressor}.

Query and document expansion have been shown to reduce vocabulary mismatch in retrieval \citep{gao2023precise, gospodinov2023doc2query}. Toollery follows a similar approach but differs in a more agentic direction - we operationalize expansion as a capability-conditioned manual construction layer rather than as generic document augmentation. Toollery does not propose a new retriever; its contribution is a deployment-oriented candidate-compression pipeline that uses generated intent manuals to bridge user requests and heterogeneous skill/tool specifications before final LLM selection.

\begin{table*}[t]
\centering
\small
\resizebox{\textwidth}{!}{%
\begin{tabular}{llccccccc}
\hline
\textbf{Retrieval Setting} &
\textbf{Retriever} &
\textbf{Online ms/query} &
\textbf{Hit@1} &
\textbf{Recall@10} &
\textbf{Recall@50} &
\textbf{MRR@10} &
\textbf{nDCG@10} &
\textbf{FullCoverage@10} \\
\hline
Raw specification & BM25 & 309.4 & 0.553 & 0.571 & 0.726 & 0.625 & 0.512 & 0.413 \\
Raw specification & RAGAnything & 38210.6 & 0.600 & 0.568 & 0.712 & 0.663 & 0.500 & 0.393 \\
Raw specification & SkillRouter$^\dagger$ & 495.8 & 0.740 & 0.704 & 0.814 & 0.791 & 0.670 & 0.520 \\
\hline
Toollery (manual only) & BM25 & \textbf{218.9} & 0.840 & 0.880 & 0.951 & 0.881 & 0.846 & 0.760 \\
Toollery (manual only) & RAGAnything & 21505.9 & 0.900 & 0.922 & 0.953 & 0.925 & \textbf{0.882} & 0.827 \\
\hline
Toollery (manual + specification) & BM25 & 623.0 & 0.920 & \textbf{0.942} & \textbf{0.968} & 0.942 & 0.880 & \textbf{0.853} \\
Toollery (manual + specification) & RAGAnything & 41070.5 & \textbf{0.933} & 0.900 & 0.950 & \textbf{0.943} & 0.843 & 0.827 \\
\hline
\end{tabular}
}
\caption{SkillRouter benchmark retrieval over roughly 79K skills, averaged over the easy and hard splits. Raw specification retrieves from original skill specifications only. Full-library prompting at this scale would require roughly 43.1M input tokens per request; Toollery top-10 prompting requires roughly 27.7K tokens, over three orders of magnitude smaller. $^\dagger$ denotes the SkillRouter reference using SR-Emb-0.6B $\times$ SR-Rank-0.6B.}
\label{tab:skillrouter-skillbench-results}
\end{table*}

\section{The Toollery Design}
\label{sec:method}

Toollery separates full-library search from final LLM decision-making. Figure~\ref{fig:toollery-overview} shows the overall pipeline: Toollery first generates user-intent queries for each selectable capability, then uses a standard retriever at inference time to retrieve a compact candidate set for the final LLM. Throughout the paper, we use \emph{capability} as the general unit of selection: a capability may be a high-level procedural skill, an MCP service, an atomic tool, or an API function. Toollery can retrieve from either generated user-intent queries alone or skill/tool specifications augmented with those generated queries; the latter preserves both user-facing intent signals and developer-facing specification details.

\subsection{Problem Formulation}
\label{subsec:problem-formulation}

Let $\mathcal{A}=\{a_1,\ldots,a_M\}$ denote the full library of available capabilities, where each capability may be a high-level skill, an MCP service, an atomic tool, or an API function. Each capability $a_i$ has a specification $s_i$, which may include its name, description, schema, instruction, usage constraints, examples, or documentation. Given a user request $x$, the goal is to construct a compact candidate set
\begin{equation}
    \mathcal{C}_k(x)\subset\mathcal{A},\qquad |\mathcal{C}_k(x)|=k\ll M,
\end{equation}
that contains the capability or capabilities needed to satisfy the request. The final LLM then receives $x$ and the specifications of only the candidates in $\mathcal{C}_k(x)$, and produces the final selection and any required arguments.

The main difficulty is the mismatch between user requests and specifications. Users describe goals, constraints, or partial tasks, while specifications describe capabilities, parameters, and usage details. Toollery addresses this mismatch by constructing a manual for the full library: a set of generated user-intent queries that describe when each capability should be selected.

\subsection{Manual Construction}
\label{subsec:intent-index}

For each capability $a_i$, a generator model $G$ produces $N$ user-intent queries from its specification:
\begin{equation}
    Q_i = G(s_i; N, \pi),
\end{equation}
where $\pi$ prompts the generator to write realistic user requests that would require $a_i$. The manual for the full library is the union of the generated user-intent queries for all capabilities:
\begin{equation}
    \mathcal{M}=\{(q,a_i): q\in Q_i,\ a_i\in\mathcal{A}\}.
\end{equation}
Each manual entry stores a generated user-intent query and the identifier of the capability whose specification produced it. We keep specifications and generated queries as separate provenance-linked fields; the final LLM receives only the authoritative specifications. Unless otherwise stated, we generate $N=3$ queries per capability. Generation uses only capability specifications and metadata, finishes before test retrieval, and receives no evaluation requests, demonstrations, or labels.


Toollery optionally applies an offline source-capability recovery filter to improve manual quality. For each generated user-intent query, a verifier is asked to identify the source capability from a pool containing that capability and eight randomly sampled distractors. Toollery keeps the query only when the verifier recovers the source capability, filtering out intents that are too broad or non-discriminative. This step uses no evaluation requests or gold labels, so it acts as an offline quality-control step for generated intents rather than a benchmark-specific reranker. Appendix~\ref{app:reproducibility} provides implementation details, Appendix~\ref{app:manual-construction-prompt} gives the prompt template, and Appendix~\ref{app:pipeline-walkthrough} shows a concrete walkthrough.

\subsection{Retrieval over the Manual}
\label{subsec:retrieval-aggregation}

At inference time, Toollery retrieves candidates using the user request $x$ and the manual $\mathcal{M}$. The retriever is not tied to a specific implementation: it can be instantiated with sparse lexical retrieval such as BM25, dense retrieval, or a RAG-style retriever. In this work, we provide two Toollery retrieval settings. The first setting scores generated user-intent queries in the manual directly, and the second setting scores specifications augmented with the corresponding user-intent queries from the manual. 

When retrieval is performed over manual entries, multiple retrieved user-intent queries may point to the same capability. Toollery therefore aggregates query-level evidence into capability-level scores. Let $f(x,q)$ be the retrieval score between user request $x$ and a generated user-intent query $q$. Toollery ranks each capability by aggregating the strongest matching queries associated with that capability:
\begin{equation}
    S(x,a_i)=\operatorname{Agg}\{f(x,q): (q,a_i)\in\mathcal{M}\}.
\end{equation}
The candidate set is then
\begin{equation}
    \mathcal{C}_k(x)=\operatorname{TopK}_{a_i\in\mathcal{A}} S(x,a_i).
\end{equation}
In our implementation, $\operatorname{Agg}$ averages the strongest retrieved query scores for each capability and adds a small evidence-count bonus; exact aggregation parameters are listed in Appendix~\ref{app:reproducibility}.

\subsection{Candidate Compression and Final Selection}
\label{subsec:selection}

For each selected capability, Toollery keeps only the specification fields required for the final LLM decision. The final LLM receives the original user request and the compact specifications for $\mathcal{C}_k(x)$, rather than the full library or the manual; retrieval narrows the action space, while the LLM resolves residual ambiguity, selects the final capability, and fills arguments.

This design changes the online cost from full-library prompting to subset prompting. Let $M$ be the number of capabilities and $L_s$ the average compact-specification length. Full-library prompting sends approximately $O(M\cdot L_s)$ specification tokens, so prompt cost grows linearly with the library. Toollery instead sends only the retrieved top-$k$ specifications, approximately $O(k\cdot L_s)$ tokens, where $k\ll M$ is a deployment-time candidate budget. Because $k$ can remain fixed as $M$ grows, the final LLM sees a bounded candidate set while the manual and retrieval index absorb library growth offline.

\begin{table*}[t]
\centering
\small
\resizebox{\textwidth}{!}{%
\begin{tabular}{llccccccc}
\hline
\textbf{Retrieval Setting} &
\textbf{Retriever} &
\textbf{Online ms/query} &
\textbf{Hit@1} &
\textbf{Recall@10} &
\textbf{Recall@50} &
\textbf{MRR@10} &
\textbf{nDCG@10} &
\textbf{FullCoverage@10} \\
\hline
Raw specification & BM25 & 3.6 & 0.643 & 0.931 & 0.976 & 0.744 & 0.790 & 0.931 \\
Raw specification & RAGAnything & 152.5 & 0.737 & 0.976 & 0.995 & 0.831 & 0.867 & 0.976 \\
\hline
Toollery (manual only) & BM25 & \textbf{0.3} & 0.468 & 0.822 & 0.934 & 0.584 & 0.642 & 0.822 \\
Toollery (manual only) & RAGAnything & 81.8 & 0.543 & 0.855 & 0.929 & 0.653 & 0.702 & 0.855 \\
\hline
Toollery (manual + specification) & BM25 & 5.0 & 0.731 & 0.948 & 0.986 & 0.811 & 0.845 & 0.948 \\
Toollery (manual + specification) & RAGAnything & 139.7 & \textbf{0.776} & \textbf{0.980} & \textbf{0.997} & \textbf{0.856} & \textbf{0.887} & \textbf{0.980} \\
\hline
\end{tabular}
}
\caption{BFCL-V4 live multiple retrieval over the complete 457-tool pool.}
\label{tab:atomic-tool-results}
\end{table*}

\begin{table*}[t]
\centering
\small
\resizebox{\textwidth}{!}{%
\begin{tabular}{llccccccc}
\hline
\textbf{Retrieval Setting} &
\textbf{Retriever} &
\textbf{Online ms/query} &
\textbf{Hit@1} &
\textbf{Recall@10} &
\textbf{Recall@50} &
\textbf{MRR@10} &
\textbf{nDCG@10} &
\textbf{FullCoverage@10} \\
\hline
Raw specification & BM25 & 14.4 & 0.780 & 0.970 & 0.995 & 0.854 & 0.883 & 0.970 \\
Raw specification & RAGAnything & 524.6 & 0.825 & \textbf{1.000} & \textbf{1.000} & 0.895 & 0.922 & \textbf{1.000} \\
\hline
Toollery (manual only) & BM25 & \textbf{0.2} & 0.555 & 0.885 & 0.990 & 0.668 & 0.721 & 0.885 \\
Toollery (manual only) & RAGAnything & 77.9 & 0.605 & 0.925 & 0.985 & 0.722 & 0.772 & 0.925 \\
\hline
Toollery (manual + specification) & BM25 & 25.9 & 0.815 & \textbf{1.000} & \textbf{1.000} & 0.887 & 0.915 & \textbf{1.000} \\
Toollery (manual + specification) & RAGAnything & 657.3 & \textbf{0.845} & \textbf{1.000} & \textbf{1.000} & \textbf{0.904} & \textbf{0.928} & \textbf{1.000} \\
\hline
\end{tabular}
}
\caption{BFCL-V4 non-live multiple retrieval over the complete 443-tool pool.}
\label{tab:atomic-tool-results-bfcl-nonlive}
\end{table*}

\begin{table}[t]
\centering
\small
\resizebox{\columnwidth}{!}{%
\begin{tabular}{llcccc}
\hline
\textbf{Dataset} &
\textbf{Selection Setting} &
\textbf{\# Candidates} &
\textbf{Input Tokens} &
\textbf{Latency} &
\textbf{Accuracy} \\
\hline
BFCL live & Full-library prompt & 457 & 84.1K & 11.8s & .770 \\
BFCL live & Raw specification top-10 & 10 & 2.7K & 2.8s & .834 \\
BFCL live & Toollery top-10 & 10 & 2.1K & 1.6s & \textbf{.837} \\
\hline
BFCL non-live & Full-library prompt & 443 & 57.0K & 8.4s & .703 \\
BFCL non-live & Raw specification top-10 & 10 & 1.9K & 2.2s & .825 \\
BFCL non-live & Toollery top-10 & 10 & 1.3K & 1.5s & \textbf{.830} \\
\hline
Cockpit & Raw specification top-10 & 10 & 2.2K & 3.5s & .163 \\
Cockpit & Toollery top-10 & 10 & 1.9K & 2.8s & \textbf{.268} \\
\hline

\end{tabular}
}
\caption{End-to-end selection under full-library prompting and fixed top-10 retrieval. The BFCL full-library rows retain the DeepSeek V3.2 results as scaling references, whereas all fixed top-10 rows use DeepSeek-V4-Pro. BFCL reports official AST Accuracy and Cockpit reports exact-set accuracy. Input tokens and latency are per-call means.}
\label{tab:atomic-tool-cost}
\end{table}

\section{Experiments}

\label{sec:experiments}

We evaluate Toollery as a candidate compression layer for scalable capability selection. The experiments test whether Toollery preserves the needed capabilities in compact top-$k$ candidate sets and whether this compression improves end-to-end LLM selection at a matched candidate budget.

\subsection{Experimental Setup}
\label{subsec:experimental-setup}

\paragraph{Benchmarks.}
We evaluate Toollery on two public benchmarks and one proprietary industrial dataset. The SkillRouter benchmark \citep{li2026skillsbench, zheng2026skillrouter} contains 75 evaluation queries: 24 single-skill queries and 51 multi-skill queries, each evaluated against a roughly 79K-capability pool. BFCL-V4 \citep{patilberkeley} provides 200 non-live multiple requests over 443 tools and 1,053 live multiple requests over 457 tools. Finally, we evaluate on 3,396 real proprietary smart-cockpit requests over a 220-tool production pool; half contain two gold tools and half contain three. This last setting measures unordered multi-intent set coverage, not dependent multi-step execution.

\paragraph{Settings and baselines.}
We compare retrieval settings that differ in what information is available to the retriever. \emph{Raw specification} retrieves only from original specifications. Toollery evaluates \emph{manual only} and \emph{manual + specification} representations using BM25 or RAGAnything. For the direct document-expansion comparison, we independently reimplement Re-Invoke's expansion and retrieval path with ten queries per capability and disable its online intent LLM \citep{chen-etal-2024-invoke}. This is an expansion-only baseline under the same no-online-generation, top-10 serving contract, not a reproduction of the complete Re-Invoke system.

\paragraph{Metrics and cost.}

For retrieval, we report Hit@1, Recall@10/50, MRR@10, nDCG@10, FullCoverage@10, and Avg ms/query for warmed online retrieval latency. Recall is the fraction of gold capabilities retrieved; FullCoverage equals one only when every gold capability appears in the top 10. The reported values are averages over all requests. Cockpit end-to-end accuracy requires the unordered predicted tool set to equal the gold set. BFCL AST Accuracy uses the official category-specific checker for call count, names, and arguments. Retrieval misses and invalid outputs remain in every denominator.

Offline manual construction cost is reported in Appendix~\ref{app:reproducibility}. Using official DeepSeek-V4-Pro prices as of June 9, 2026,\footnote{\url{https://api-docs.deepseek.com/quick_start/pricing}, accessed June 9, 2026.} construction costs approximately \$0.22 per 100 skills on average, including generated-query construction and source-capability recovery filtering. This offline cost is excluded from the online retrieval latency numbers in the main tables.

\begin{table}[t]
\centering
\small
\setlength{\tabcolsep}{3.2pt}
\resizebox{\columnwidth}{!}{%
\begin{tabular}{lccc}
\hline
Data & Raw specification & Re-Invoke$^\dagger$ & Toollery \\
 & ($N=0$) & ($N=10$) & ($N=3$) \\
\hline
Cockpit & .545 / .224 & .593 / .240 & \textbf{.684 / .381} \\
BFCL non-live & 1.000 / 1.000 & 1.000 / 1.000 & 1.000 / 1.000 \\
BFCL live & .976 / .976 & .976 / .976 & \textbf{.980 / .980} \\
\hline
\end{tabular}
}
\caption{Retrieval coverage under the same top-10 candidate budget. Each cell reports Recall@10 / FullCoverage@10, and $N$ is the number of generated queries per capability. Cockpit uses BM25, whereas BFCL uses Qwen3-Embedding-0.6B dense retrieval. $^\dagger$Expansion-only Re-Invoke reimplementation without its online intent model.}
\label{tab:matched-retrieval}
\end{table}

\begin{table*}[t]
\centering
\small
\resizebox{\textwidth}{!}{%
\begin{tabular}{llcccccccc}
\hline
\textbf{Retrieval Setting} &
\textbf{Retriever} &
\textbf{Online ms/query} &
\textbf{Hit@1} &
\textbf{Recall@10} &
\textbf{Recall@20} &
\textbf{Recall@50} &
\textbf{MRR@10} &
\textbf{nDCG@10} &
\textbf{FullCoverage@10} \\
\hline
Raw specification & BM25 & 6.4 & 0.430 & 0.545 & 0.696 & 0.867 & 0.576 & 0.446 & 0.224 \\
Raw specification & RAGAnything & 144.5 & 0.408 & 0.582 & 0.718 & 0.871 & 0.588 & 0.464 & 0.229 \\
\hline
Toollery (manual only) & BM25 & \textbf{0.7} & 0.481 & 0.670 & 0.784 & 0.918 & 0.648 & 0.544 & 0.358 \\
Toollery (manual only) & RAGAnything & 221.6 & 0.453 & 0.666 & 0.781 & 0.902 & 0.633 & 0.538 & 0.357 \\
\hline
Toollery (manual + specification) & BM25 & 69.7 & 0.484 & \textbf{0.684} & \textbf{0.803} & \textbf{0.926} & 0.654 & \textbf{0.553} & \textbf{0.381} \\
Toollery (manual + specification) & RAGAnything & 253.4 & 0.403 & 0.630 & 0.758 & 0.893 & 0.591 & 0.500 & 0.316 \\
\hline
\end{tabular}
}

\caption{Proprietary smart cockpit retrieval over the complete 220-tool pool. Results are averaged over 3,396 real company-internal requests with two or three gold tools. The proprietary dataset is used only for deployment-facing evaluation and is not released.}
\label{tab:internal-car-results}
\end{table*}

\subsection{Skill Retrieval on the SkillRouter Benchmark}
\label{subsec:skill-routing-results}

The SkillRouter benchmark tests Toollery at a scale where full-library prompting is infeasible. Prompting over the full \textasciitilde{}79K-capability library would require roughly 43M input tokens per request, while Toollery top-10 prompting requires about 27.7K tokens. The question is therefore whether retrieval can compress this full library into a small candidate set without losing the needed capabilities.

Table~\ref{tab:skillrouter-skillbench-results} shows that Toollery substantially improves large-library retrieval. With BM25, Toollery (manual + specification) improves Hit@1 from 0.553 to 0.920, Recall@10 from 0.571 to 0.942, and FullCoverage@10 from 0.413 to 0.853 over raw specification retrieval. The paired FullCoverage improvement is 0.440 (95\% CI [0.340, 0.547]) across 75 requests. The table's latency column reports retrieval time only; with BM25, Toollery performs this \textasciitilde{}79K-capability compression in 623.0 ms/query. 

\subsection{Atomic-Tool Retrieval and End-to-End Selection on BFCL-V4}
\label{subsec:atomic-retrieval-results}

BFCL-V4 tests whether the same manual-based design transfers from high-level skills to atomic tools and API functions. Unlike SkillRouter, BFCL-V4 is small enough to use full-library prompting as a scaling reference, but our primary quality comparison holds the retrieved candidate budget at ten. Tables~\ref{tab:atomic-tool-results} and~\ref{tab:atomic-tool-results-bfcl-nonlive} evaluate retrieval over the complete BFCL-live 457-tool pool and BFCL non-live 443-tool pool.

Toollery remains effective in this atomic-tool setting. With RAGAnything, Toollery (manual + specification) reaches 0.776 Hit@1 and 0.980 Recall@10 on BFCL-live, and 0.845 Hit@1 and 1.000 Recall@10 on BFCL non-live. The manual-only setting slightly underperforms the raw specification retrieval because the atomic tool specs are already short and schema-structured,  but the manual still improves top-$k$ recall by adding user-intent signals; adding specifications back recovers and exceeds the baseline.

Table~\ref{tab:matched-retrieval} places these results under a matched serving contract. Raw dense retrieval, Re-Invoke expansion, and Toollery all reach 1.000 Recall and FullCoverage on non-live; on live, Toollery obtains 0.980 versus 0.976 for both baselines. Table~\ref{tab:atomic-tool-cost} retains the DeepSeek V3.2 full-library results as scaling references and compares raw-specification retrieval with Toollery while holding the candidate budget, final selector, prompt, decoding, and evaluation population fixed. BFCL AST Accuracy changes from 0.825 to 0.830 on multiple and from 0.834 to 0.837 on live multiple; neither difference is statistically significant. Toollery therefore provides little quality gain once raw top-10 retrieval is saturated. Additional end-to-end scaling results appear in Appendix~\ref{app:end-to-end-results}.

\subsection{Atomic-Tool Retrieval on Proprietary Dataset}
\label{subsec:internal-car-results}

The proprietary smart cockpit dataset evaluates Toollery in a production-style multi-intent setting, where each real user request maps to two or three gold tools from a 220-tool pool. Full-library prompting over this pool would require roughly 35K input tokens per request; Toollery reduces this to about 1.9K tokens with top-10 candidates and about 9.1K tokens with top-50 candidates.

Table~\ref{tab:internal-car-results} reports the complete cockpit retrieval results, while Table~\ref{tab:matched-retrieval} compares raw specification retrieval, expansion-only Re-Invoke, and Toollery under the same top-10 candidate budget. Compared with raw BM25 specification retrieval, Toollery improves Recall@10 from 0.545 to 0.684 and FullCoverage@10 from 0.224 to 0.381. Against expansion-only Re-Invoke, the gains are 0.091 Recall (95\% CI [0.080, 0.102]) and 0.141 FullCoverage ([0.123, 0.158]), despite using three rather than ten generated queries. Fixed-budget exact tool-set selection rises from 0.163 to 0.268, a 0.105 gain ([0.092, 0.119]). The end-to-end gain accompanies an increase in FullCoverage@10 from 0.224 to 0.381.

\section{Discussion and Conclusion}
\label{sec:discussion_conclusion}

The experiments show that full-library prompting quickly becomes a poor serving strategy as capability libraries grow. At SkillRouter scale, it is infeasible: prompting over the full \textasciitilde{}79K-capability library would require roughly 43.1M input tokens per request. Even at smaller scales, the cost remains substantial: BFCL-V4 requires 57.0K--84.1K input tokens and 8.4--11.8 seconds of latency over 443--457 tools, while the 220-tool smart cockpit pool still requires roughly 35.2K input tokens. The deployment pressure is therefore structural: the final LLM pays for every available capability on every request, including trivial or irrelevant ones.

Toollery addresses this bottleneck by moving semantic matching out of the final prompt and into offline manual construction plus online retrieval. Across all three settings, Toollery keeps final selection bounded to a compact top-$k$ candidate set: SkillRouter drops from roughly 43.1M full-library tokens to 27.7K top-10 tokens; BFCL-V4 drops from 57.0K--84.1K to 1.3K--2.1K tokens; and the proprietary dataset drops from 35.2K to 1.9K top-10 tokens. On BFCL-V4, these reductions also translate into latency drops from 8.4--11.8s to 1.5--1.6s.

The accuracy results show that Toollery does not buy efficiency at the expense of selection quality. Manual-based retrieval improves or matches raw specification retrieval across SkillRouter, BFCL-V4, and the proprietary multi-intent dataset, while substantially improving candidate-set coverage for multi-capability requests. On BFCL-V4, where raw top-10 dense retrieval is already near saturation, Toollery maintains comparable fixed-budget AST Accuracy while substantially reducing prompt context and latency. Together, these results show that candidate compression can improve coverage where retrieval is limiting without sacrificing end-to-end selection quality in saturated settings.

Taken together, these results suggest that scalable capability selection should be treated as candidate compression rather than prompting over an ever-growing full library. Toollery instantiates this principle by constructing a provenance-linked manual offline, retrieving a compact candidate set online, and allowing the final LLM to decide only over authoritative retrieved specifications. Overall, Toollery provides a practical retrieval-and-selection layer when paired with workload-specific calibration and fallback behavior.

\newpage
\section{Limitations}
\label{sec:limitations}

Toollery depends on the coverage and precision of the manual. If the generated user-intent queries fail to cover realistic ways users ask for a capability, retrieval recall is bounded regardless of the retrieval backend. Conversely, overly broad user-intent queries can introduce noisy matches and increase distractor exposure in the retrieved candidate set. Source-capability recovery filtering reduces this risk by removing queries that do not recover their source capability under distractor competition, but it does not provide a formal correctness guarantee. As shown in Appendix~\ref{app:failure-analysis}, remaining failures often involve functionally similar capabilities or mismatches between broad retrieved capabilities and finer-grained benchmark gold labels; exact recall may therefore under-credit plausible alternatives while still exposing real disambiguation challenges.

Toollery also introduces an offline maintenance cost. When capabilities are added, removed, or modified, the corresponding manual entries and retrieval index must be updated. This cost is amortized across future requests and separated from online retrieval latency, but it may matter for rapidly changing capability libraries. Our experiments cover large public skill/tool pools and one proprietary smart cockpit deployment setting; broader validation across more domains, dynamic capability updates, and interactive multi-step agent workflows remains future work.

\bibliography{custom}

\clearpage
\newpage
\appendix

\section{Reproducibility and Deployment Details}
\label{app:reproducibility}

Table~\ref{tab:reproducibility-details} summarizes the main configuration choices used in the reported runs. We separate offline manual construction from online retrieval and final selection: manual construction includes user-intent query generation and optional source-capability recovery filtering, while online latency reports only serving-time retrieval or final-selection latency with warmed indexes and caches. We also plan to provide plugin adapters for popular agent runtimes, including OpenClaw, Hermes Agent, Claude Code, and Codex, so that Toollery can be used as a candidate-compression layer in existing agent systems.

\begin{table*}[t]
\centering
\footnotesize
\setlength{\tabcolsep}{4pt}
\renewcommand{\arraystretch}{1.08}
\begin{tabular}{p{0.24\textwidth}p{0.70\textwidth}}
\hline
\textbf{Item} & \textbf{Value used in reported runs} \\
\hline
SkillRouter pools & 78,361 easy capabilities; 79,141 hard capabilities; 75 requests per pool \\
BFCL-live end-to-end pool & 457 tools; 1,053 live multiple requests; final column uses the complete pool \\
BFCL-non-live end-to-end pool & 443 tools; 200 multiple requests; final column uses the complete pool \\

Manual construction & DeepSeek-V4-Pro through an OpenAI-compatible API; 3 user-intent queries per capability; seed 31; 4 workers; batch size 4 \\
Source-capability recovery verifier & asks DeepSeek-V4-Pro to choose the matching capability id from the source capability plus distractors, or NONE if no candidate fits \\

Distractor construction & verifier pool uses the source capability plus 8 randomly sampled non-gold distractors; scaled BFCL candidate pools sample the gold tool plus non-gold tools without replacement \\
BM25 configuration & local tokenizer in \texttt{toollery.text}; Okapi BM25 with $k_1=1.5,b=0.75$; Toollery (manual + specification) concatenates name, category, description, schema, and user-intent queries from the manual \\
RAGAnything configuration & text-only RAGAnything/LightRAG cache with one compact chunk per capability; retrieval queries LightRAG chunk vectors and recovers capability ids from returned context order \\
Retrieval embedding model & Qwen/Qwen3-Embedding-0.6B for neural retrieval and latency replay \\
Aggregation parameters & top-$c=3$ averaged manual-query scores with evidence bonus $b=5,\lambda=0.05$ \\
Candidate budgets & top-50 for retrieval-quality tables; top-10 for the main BFCL-V4 end-to-end cost comparison; appendix end-to-end table columns denote the candidate-pool size before Toollery compression \\
Latency protocol & warmed indexes/caches; reports online retrieval or final-selection latency only; excludes offline manual construction, verification, embedding, and index construction \\
Offline cost & SkillRouter easy and hard pools; DeepSeek-V4-Pro generator and verifier; 3 user-intent queries per capability; source capability plus 8 random distractors for verification; cost computed with cache-miss input pricing \\

Release plan & Upon acceptance, we will release the Toollery code, prompts, normalized manual artifacts, run configurations, metric scripts, and plugin adapters publicly. \\
\hline
\end{tabular}
\caption{Reproducibility details for the main retrieval and efficiency results. Secrets and local API endpoints are omitted; paths in the released run configurations identify the exact output artifacts.}
\label{tab:reproducibility-details}
\end{table*}

\subsection{Manual Construction Prompt}
\label{app:manual-construction-prompt}

The following prompt template is used to generate user-intent queries for manual construction in the SkillRouter and BFCL-V4 runs. Each capability is serialized by its specification, including available fields such as name, description, category, schema, and usage instructions. The generated \texttt{query} fields are stored as manual entries associated with the source capability.

\begin{center}
\fbox{
\begin{minipage}{0.94\linewidth}
\footnotesize
\ttfamily
\raggedright
Generate realistic user-intent queries for multiple skills or tools.\par
For each skill or tool, write things real users might ask when they need that exact skill or tool.\par
Use practical task context, constraints, and natural wording.\par
Avoid copying skill text verbatim.\par
Return only one valid JSON object. The keys must be the exact skill names.\par
Each value must be a JSON array of objects. Each object must contain query, scenario\_type, generation\_notes.\par
Return exactly \{N\} objects per skill.\par

\end{minipage}
}
\end{center}

The returned \texttt{query} fields become manual entries associated with the source capability, while \texttt{scenario\_type} and \texttt{generation\_notes} are retained for auditability but are not used by the retriever.

\begin{figure}[t]
    \centering
    \includegraphics[width=0.86\columnwidth]{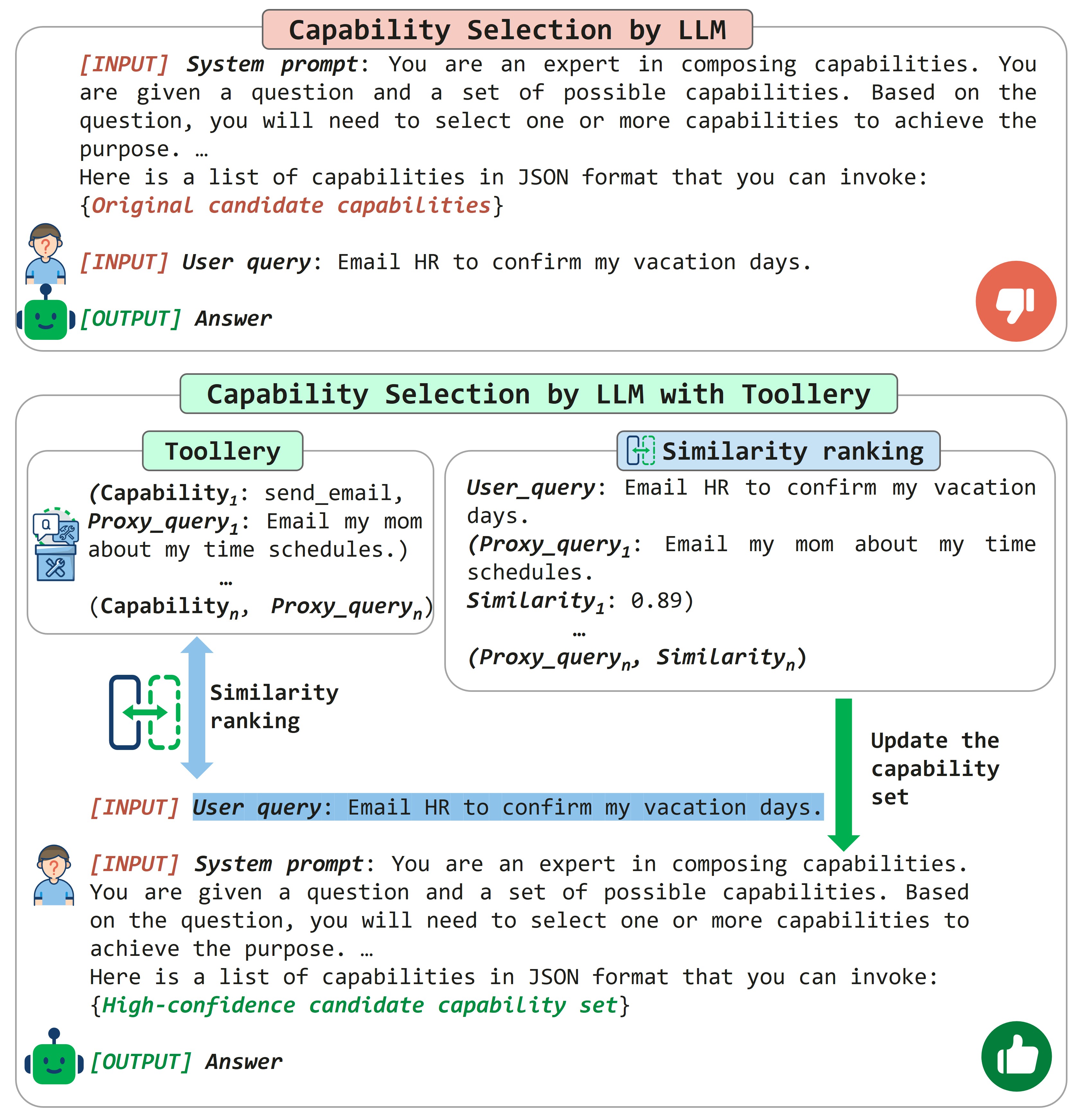}
    \caption{\textbf{Toollery pipeline walkthrough.} Toollery converts capability specifications into user-intent queries during offline manual construction. During inference, the user request retrieves matching manual entries and corresponding capability specifications, which are aggregated into a compact candidate set for final LLM decision making.}
    \label{fig:toollery-walkthrough}
\end{figure}

\subsection{Source-Capability Recovery Filtering}
\label{app:source-capability-filtering}

For each generated user-intent query $q \in Q_i$, Toollery optionally constructs a verification pool
$\mathcal{P}_{i,q}=\{a_i\}\cup\mathcal{D}_{i,q}$, where $a_i$ is the source capability whose specification produced $q$ and $\mathcal{D}_{i,q}$ contains eight randomly sampled distractor capabilities from the same full library. A verifier LLM receives $q$ and the specifications of all capabilities in $\mathcal{P}_{i,q}$, then selects the best matching capability or abstains. Toollery keeps $q$ only when the verifier recovers the source capability $a_i$.

This filter removes user-intent queries that are too broad or non-discriminative before indexing. Because filtering is performed during offline manual construction, its cost is amortized across future online requests together with query generation.

\subsection{Pipeline Walkthrough Example}
\label{app:pipeline-walkthrough}

Figure~\ref{fig:toollery-walkthrough} illustrates one concrete Toollery workflow. Starting from a capability specification, Toollery first generates user-intent queries that describe realistic requests for which the capability should be selected. These generated queries are stored in the manual and remain linked to the source capability. At inference time, a real user request is matched against the manual and, optionally, the original capability specifications. Retrieved manual entries are aggregated back to capability-level scores, producing a compact top-$k$ candidate set. The final LLM then receives only the user request and the retrieved candidate specifications, rather than the full library, and performs the final schema-aware decision. This example highlights the main serving-time change: the final LLM no longer pays the token, latency, and distractor cost of the full library, but only reasons over a bounded set of retrieved candidate specifications.

\subsection{Offline Manual Construction Cost}
\label{app:offline-construction-cost}

We report offline manual construction cost separately from online latency because this cost is paid before deployment and amortized across future requests. On SkillRouter, we use DeepSeek-V4-Pro as both the generator and the source-capability recovery verifier. Each capability receives three generated user-intent queries, and each non-empty query is verified against its source capability plus eight randomly sampled distractors.

Table~\ref{tab:offline-cost-sample} reports the measured construction cost for the SkillRouter easy and hard pools. Under official DeepSeek-V4-Pro cache-miss input and output pricing,\footnote{\url{https://api-docs.deepseek.com/quick_start/pricing/}, accessed June 9, 2026.} offline manual construction costs approximately \$0.22 per 100 capabilities on average. These values describe offline construction only; all online latency results in the main tables use warmed indexes and exclude generation, verification, embedding, and index construction.

\begin{table}[t]
\centering
\scriptsize
\setlength{\tabcolsep}{2.5pt}
\begin{tabular}{lrrrrr}
\hline
Setting & Skills & Cand. & Ret. & Tok. & Cost \\
\hline
SkillRouter easy & 78,361 & 235K & 228K & 361.8M & \$168.70 \\
SkillRouter hard & 79,141 & 237K & 230K & 365.4M & \$170.38 \\
\hline
\end{tabular}
\caption{Offline construction cost for generated-query construction and verifier filtering. The measured rows use the SkillRouter easy and hard pools with three generated queries per capability. Cost uses official DeepSeek-V4-Pro pricing: \$0.435/M cache-miss input tokens and \$0.87/M output tokens; cache-hit discounts are not applied.}

\label{tab:offline-cost-sample}
\end{table}

\section{Additional Analyses and Deployment Stress Tests}
\label{app:additional-analyses}

Table~\ref{tab:representation-retriever} isolates the interaction between indexed representation and retriever at fixed $N=3$, $k=10$, and random-negative recovery. Generated queries only is a retrieval ablation: Toollery still loads authoritative specifications before final selection. Dense retrieval uses Qwen3-Embedding-0.6B, and hybrid retrieval uses equal-weight reciprocal-rank fusion with RRF $k=60$.

\begin{table*}[t]
\centering
\small
\setlength{\tabcolsep}{4.2pt}
\renewcommand{\arraystretch}{1.10}
\resizebox{\textwidth}{!}{%
\begin{tabular}{llrrrrrr}
\toprule
\multirow{2}{*}{\textbf{Dataset}} &
\multirow{2}{*}{\textbf{Indexed representation}} &
\multicolumn{2}{c}{\textbf{BM25}} &
\multicolumn{2}{c}{\textbf{Dense (Qwen3)}} &
\multicolumn{2}{c}{\textbf{Hybrid RRF}} \\
\cmidrule(lr){3-4}\cmidrule(lr){5-6}\cmidrule(lr){7-8}
 & & \textbf{R@10} & \textbf{FC@10} & \textbf{R@10} & \textbf{FC@10} & \textbf{R@10} & \textbf{FC@10} \\
\midrule
\multirow{3}{*}{\textbf{BFCL non-live}}
& Specification & 0.970 & 0.970 & \textbf{1.000} & \textbf{1.000} & 0.990 & 0.990 \\
& Generated queries & 0.885 & 0.885 & 0.925 & 0.925 & 0.975 & 0.975 \\
& Specification + generated queries & \textbf{1.000} & \textbf{1.000} & \textbf{1.000} & \textbf{1.000} & \textbf{1.000} & \textbf{1.000} \\
\midrule
\multirow{3}{*}{\textbf{BFCL live}}
& Specification & 0.931 & 0.931 & 0.976 & 0.976 & 0.976 & 0.976 \\
& Generated queries & 0.822 & 0.822 & 0.855 & 0.855 & 0.916 & 0.916 \\
& Specification + generated queries & \textbf{0.948} & \textbf{0.948} & \textbf{0.980} & \textbf{0.980} & \textbf{0.980} & \textbf{0.980} \\
\midrule
\multirow{3}{*}{\textbf{Cockpit}}
& Specification & 0.545 & 0.224 & 0.582 & 0.229 & 0.687 & 0.371 \\
& Generated queries & 0.670 & 0.358 & \textbf{0.666} & \textbf{0.357} & 0.631 & 0.310 \\
& Specification + generated queries & \textbf{0.684} & \textbf{0.381} & 0.630 & 0.316 & \textbf{0.709} & \textbf{0.405} \\
\midrule
\multirow{3}{*}{\textbf{SkillRouter}}
& Specification & 0.571 & 0.413 & 0.568 & 0.393 & 0.621 & 0.467 \\
& Generated queries & 0.880 & 0.760 & \textbf{0.922} & \textbf{0.827} & 0.941 & 0.840 \\
& Specification + generated queries & \textbf{0.942} & \textbf{0.853} & 0.900 & \textbf{0.827} & \textbf{0.945} & \textbf{0.867} \\
\bottomrule
\end{tabular}
}
\caption{Retrieval ablation across indexed representations and retrievers. Results report Recall@10 (R@10) and FullCoverage@10 (FC@10); bold marks the best result in each dataset--retriever--metric group, including ties. SkillRouter averages the easy and hard pools, and Hybrid RRF fuses BM25 and dense rankings.}
\label{tab:representation-retriever}
\end{table*}

No representation or retriever is uniformly best. Generated queries alone are weaker than specifications on BFCL, whereas the combined representation reaches 1.000/1.000 with BM25 and dense retrieval on BFCL non-live. On SkillRouter, generated queries improve substantially over specifications and hybrid retrieval is strongest at 0.945/0.867. On cockpit, dense retrieval improves the specification-only representation, BM25 is stronger for the two query-augmented representations, and hybrid retrieval is strongest for the combined representation at 0.709/0.405. Dense and hybrid retrieval add roughly 10 ms over BM25 in our local runs, so the representation and retriever should be selected on held-out in-domain traffic rather than treated as universally optimal.

\begin{table}[t]
\centering
\small
\setlength{\tabcolsep}{4.5pt}
\renewcommand{\arraystretch}{1.10}
\begin{tabular*}{\columnwidth}{@{\extracolsep{\fill}}lrrrr@{}}
\toprule
\textbf{Dataset} & \textbf{$N$} & \textbf{R@10} & \textbf{FC@10} & \makecell{\textbf{Query-text}\\\textbf{tokens (est.)}} \\

\midrule
\multirow{4}{*}{\textbf{BFCL non-live}}
& 1 & 0.980 & 0.980 & 10,835 \\
& \textbf{3} & \textbf{0.995} & \textbf{0.995} & \textbf{33,582} \\
& 5 & 0.995 & 0.995 & 56,012 \\
& 10 & 0.995 & 0.995 & 111,042 \\
\midrule
\multirow{4}{*}{\textbf{BFCL live}}
& 1 & 0.932 & 0.932 & 11,829 \\
& \textbf{3} & \textbf{0.942} & \textbf{0.942} & \textbf{36,078} \\
& 5 & 0.949 & 0.949 & 60,189 \\
& 10 & 0.950 & 0.950 & 121,314 \\
\midrule
\multirow{4}{*}{\textbf{Cockpit}}
& 1 & 0.585 & 0.263 & 707 \\
& \textbf{3} & \textbf{0.680} & \textbf{0.365} & \textbf{2,458} \\
& 5 & 0.691 & 0.371 & 4,310 \\
& 10 & 0.708 & 0.391 & 8,921 \\
\bottomrule
\end{tabular*}
\caption{Generated-query budget sweep at $k=10$. Bold marks the default $N=3$ configuration. Rows use nested prefixes from a separately generated $N_{\max}=10$ batch; this $N=3$ arm is therefore distinct from the main-run artifact.}
\label{tab:n-sweep}
\end{table}

BFCL non-live saturates at $N=3$. On BFCL live, FullCoverage increases from 0.932 at $N=1$ to 0.942 at $N=3$ and 0.950 at $N=10$. Cockpit gains most from the first three queries (0.263 to 0.365); increasing to $N=10$ reaches 0.391, an additional 0.026 while using 3.63$\times$ the query-text index size. We therefore use $N=3$ as a compact construction-budget choice, not a universal optimum. Across the nested prefixes, normalized exact duplicates are zero; this check does not detect fuzzy or semantic near-duplicates.

\begin{table}[t]
\centering
\small
\setlength{\tabcolsep}{2.6pt}
\renewcommand{\arraystretch}{1.12}
\begin{tabular*}{\columnwidth}{@{\extracolsep{\fill}}lccc@{}}
\toprule
\textbf{Filter} & \textbf{BFCL-NL} & \textbf{BFCL-L} & \textbf{Cockpit} \\
\midrule
No filter & \textbf{0.995 / 0.995} & 0.941 / 0.941 & 0.623 / 0.310 \\
Random neg. & \textbf{0.995 / 0.995} & \textbf{0.942 / 0.942} & \textbf{0.680 / 0.365} \\
BM25 hard neg. & \textbf{0.995 / 0.995} & 0.941 / 0.941 & 0.602 / 0.288 \\
\bottomrule
\end{tabular*}
\caption{Recovery-filter ablation at $N=3,k=10$ with eight verifier negatives. Cells report Recall@10 / FullCoverage@10; bold marks the best result per dataset, including ties. BFCL-NL and BFCL-L denote non-live and live. Hard-negative verification is diagnostic only.}
\label{tab:filter-sweep}
\end{table}

Within the separately generated $N_{\max}=10$ batch, random-negative recovery is neutral on BFCL non-live and produces the strongest $N=3$ result on BFCL live and cockpit. The largest practical gain is on cockpit, where Recall increases from 0.623 to 0.680 and FullCoverage from 0.310 to 0.365. Hard-negative recovery does not improve retrieval and reduces cockpit coverage, so we treat it as a diagnostic stress test rather than a validated improvement.

\begin{table}[t]
\centering
\small
\setlength{\tabcolsep}{5pt}
\renewcommand{\arraystretch}{1.10}
\begin{tabular*}{\columnwidth}{@{\extracolsep{\fill}}rrrr@{}}
\toprule
\textbf{$k$} & \textbf{Recall / FC} & \textbf{Candidate tokens} & \textbf{ms/query} \\
\midrule
1 & 0.700 & 254 & 0.15 \\
3 & 0.864 & 664 & 0.15 \\
5 & 0.911 & 1,063 & 0.15 \\
\textbf{10} & \textbf{0.942} & \textbf{2,039} & \textbf{0.15} \\
20 & 0.962 & 3,965 & 0.15 \\
50 & 0.979 & 9,734 & 0.17 \\
\bottomrule
\end{tabular*}
\caption{Candidate-budget sweep on BFCL live using combined BM25 retrieval. Bold marks the default $k=10$ configuration. Each request has one gold function, so Recall equals FullCoverage.}
\label{tab:k-sweep}
\end{table}

The nested generated-query batches contain no normalized exact duplicates. This check does not establish semantic diversity, so paraphrases and near-duplicates may remain.

\begin{table}[t]
\centering
\small
\setlength{\tabcolsep}{6pt}
\renewcommand{\arraystretch}{1.10}
\begin{tabular*}{\columnwidth}{@{\extracolsep{\fill}}lrr@{}}
\toprule
\textbf{Cockpit route ($k=10$)} & \textbf{R@10} & \textbf{FC@10} \\
\midrule
Held-out hot only & 0.080 & 0.004 \\
Raw BM25 & 0.545 & 0.224 \\
\textbf{Toollery} & \textbf{0.684} & \textbf{0.381} \\
3 hot + Toollery & 0.628 & 0.302 \\
5 hot + Toollery & 0.563 & 0.230 \\
\bottomrule
\end{tabular*}
\caption{Five-fold held-out evaluation of static popularity routing. Hot rankings are computed on four folds and evaluated on the unseen fold; bold marks the best route.}
\label{tab:hot-set}
\end{table}

Table~\ref{tab:hot-set} shows that, at $k=10$, Toollery exceeds the held-out hot-only route by 0.604 Recall and 0.377 FullCoverage. Reserving three or five slots for globally popular tools reduces FullCoverage by 0.079 and 0.151, respectively, because those slots displace request-specific candidates. A static popularity set is therefore not an effective substitute for request-conditioned retrieval on this relatively flat, multi-intent workload.

\subsection{Test Isolation}
\label{app:test-isolation}

Manual generation finishes before test retrieval and uses only capability specifications and metadata; no evaluation request, demonstration, or label is supplied. Table~\ref{tab:overlap} reports a post-hoc overlap audit between evaluation requests and generated queries.

\begin{table}[t]
\centering
\scriptsize
\setlength{\tabcolsep}{3.5pt}
\renewcommand{\arraystretch}{1.05}
\resizebox{\columnwidth}{!}{%
\begin{tabular}{lrrrrl}
\toprule
\textbf{Split} & \textbf{Requests} & \textbf{Capabilities} & \textbf{Exact} & \textbf{Max 5-gram Jaccard} & \textbf{Unit} \\
\midrule
SkillRouter easy & 75 & 78,361 & 0 & 0.063 & token \\
SkillRouter hard & 75 & 79,141 & 0 & 0.060 & token \\
BFCL non-live & 200 & 443 & 0 & 0.455 & token \\
BFCL live & 1,053 & 457 & 0 & 0.229 & token \\
Cockpit & 3,396 & 220 & 0 & 0.417 & character \\
\bottomrule
\end{tabular}}
\caption{Evaluation-to-generated-query overlap audit. Exact denotes normalized exact matches; English splits use token 5-grams and cockpit uses character 5-grams.}
\label{tab:overlap}
\end{table}

All splits have zero normalized exact matches. On cockpit, the maximum Jaccard pair matches one capability mentioned within a multi-intent request rather than reproducing the full request. The data-flow boundary is the primary isolation evidence; the overlap audit provides evidence against exact copying but does not establish semantic independence or rule out benchmark exposure during generator pretraining.

\subsection{Cache-Aware Replay}
\label{app:cache-replay}

\begin{table*}[t]
\centering
\small
\setlength{\tabcolsep}{3.4pt}
\renewcommand{\arraystretch}{1.08}
\resizebox{\textwidth}{!}{%
\begin{tabular}{llrrrrrrr}
\toprule
\textbf{Cache regime} & \textbf{Method} & \textbf{Calls} & \textbf{Candidates} & \textbf{AST} & \textbf{Latency} & \textbf{Billed input cost} & \textbf{Input tokens} & \textbf{Hit rate} \\
\midrule
Natural replay & Full library & 600 & 443--457 & 0.787 & 3.76\,s & \textbf{\$0.000749} & 87,074 & \textbf{0.989} \\
 & Toollery & 600 & \textbf{10} & \textbf{0.840} & \textbf{2.33\,s} & \$0.000819 & \textbf{2,196} & 0.144 \\
\midrule
Cache-suppressed & Full library & 300 & 443--457 & 0.797 & 9.03\,s & \$0.037879 & 87,078 & 0.000 \\
 & Toollery & 300 & \textbf{10} & \textbf{0.817} & \textbf{2.15\,s} & \textbf{\$0.000951} & \textbf{2,187} & 0.000 \\
\midrule
Forced-warm exact & Full library & 300 & 443--457 & 0.807 & 3.43\,s & \$0.000328 & 87,077 & \textbf{1.000} \\
 & Toollery & 300 & \textbf{10} & \textbf{0.853} & \textbf{2.14\,s} & \textbf{\$0.000036} & \textbf{2,188} & 0.970 \\
\bottomrule
\end{tabular}}
\caption{Controlled cache-aware BFCL replay. Latency, billed input cost, input tokens, and hit rate are per-request means. Forced-warm values exclude the paid priming call and represent a boundary condition rather than a deployable cost forecast.}
\label{tab:cache-replay}
\end{table*}

Table~\ref{tab:cache-replay} shows that caching changes the realized cost benefit of candidate compression. In the natural replay, Toollery reduces model-visible input by 97.5\% and mean latency by 1.43~s, but the billed-input-cost difference is inconclusive: Toollery minus full library is \$0.000070 per request (95\% CI [USD $-0.000524$, USD $0.000462$]). We therefore find no billed-input-cost advantage in this regime. With caching suppressed, Toollery reduces billed input cost by 97.5\% and latency by 6.88~s. Forced-warm exact prompts also favor Toollery, but exclude the paid priming call. Cost savings therefore depend on cache state and traffic reuse; visible-token reduction alone does not imply proportional monetary savings.

\FloatBarrier

\begin{table*}[htb]
\caption{\label{bfcl_live_Result}
    BFCL-V4 live multiple results
  }
\centering
\resizebox{\textwidth}{!}{
\begin{tabular}{lcccccccc}
\hline \hline
\textbf{Model} &
  \textbf{$\leq$ 10} & \textbf{20} & \textbf{30} & \textbf{40} & \textbf{50} & \textbf{100} & \textbf{457} \\ 
  \hline \hline
\textbf{Qwen3-4B (Prompt)}                  & 85.50 & 86.00 & 88.00 & 82.50 & 78.00  & 69.50 & 24.88 & \\
\textbf{Qwen3-4B (Prompt) + Toollery}       & 87.00 $\Uparrow$ & 84.50 & 87.00 & 86.50 $\Uparrow$ & 82.50 $\Uparrow$  & 81.00 $\Uparrow$ & 83.75 $\Uparrow$ \\ \hline
\textbf{Qwen3-8B (Prompt)}                  & 97.00 & 93.50 & 92.00 & 84.00 & 83.00  & 84.50 & 38.50 \\
\textbf{Qwen3-8B (Prompt) + Toollery}       & 95.50 & \textbf{98.50} $\Uparrow$ & \textbf{98.50} $\Uparrow$ & 92.50 $\Uparrow$ & \textbf{94.50}  & 86.50 $\Uparrow$ & 79.38 $\Uparrow$ \\ \hline
\textbf{DeepSeek V3.2  (Prompt)}            & 94.50 & 80.00 & 84.00 & 89.50 & 79.50  & 63.00 & 77.00 \\
\textbf{DeepSeek V3.2  (Prompt) + Toollery} & 98.00 $\Uparrow$ & 78.50 & 94.50 $\Uparrow$ & \textbf{99.50} $\Uparrow$ & 89.50 $\Uparrow$  & 89.00 $\Uparrow$ & 83.13 $\Uparrow$  \\ \hline
\textbf{A major proprietary LLM (Prompt)}               & 83.00 & 93.00 & 90.00 & 87.00 & 88.00  & 88.00 & 79.75            \\
\textbf{A major proprietary LLM (Prompt) + Toollery}    & 83.50 $\Uparrow$ & 94.00 $\Uparrow$ & 89.50 & 95.00 $\Uparrow$ & 89.50 $\Uparrow$  & 89.00 $\Uparrow$ & 84.50 $\Uparrow$            \\ \hline
\textbf{A major proprietary LLM (FC)}                    & 67.00 & 62.00 & 63.50 & 62.00 & 61.00  & 63.50 & 57.00            \\
\textbf{A major proprietary LLM (FC) + Toollery}         & 74.50 $\Uparrow$ & 67.50 $\Uparrow$ & 69.00  $\Uparrow$& 69.50 $\Uparrow$ & 67.00 $\Uparrow$  & 67.50 $\Uparrow$ & 66.00 $\Uparrow$            \\ \hline
\textbf{Qwen3-8B (FC)}                      & 93.50 & 94.50 & 92.50 & 83.00 & 83.50  & 83.00 & 60.00 \\
\textbf{Qwen3-8B (FC) + Toollery}           & 93.50 & 98.00 $\Uparrow$ & 90.50 & 99.00 $\Uparrow$ & 91.50 $\Uparrow$  & 88.50 $\Uparrow$ & 85.88 $\Uparrow$  \\ \hline
\textbf{Qwen3-235B (FC)}                    & 98.00 & 93.00 & 87.50 & 89.50 & 85.50 & 80.00 & 81.13            \\
\textbf{Qwen3-235B (FC) + Toollery}         & \textbf{98.50} $\Uparrow$ & 92.50 & 89.00 $\Uparrow$ & 89.50 & 90.00  & \textbf{91.50} $\Uparrow$ & \textbf{88.63} $\Uparrow$ \\ \hline \hline
\end{tabular}
}
\vskip -0.1in
\end{table*}
\begin{table*}[htb]
\caption{\label{bfclResult-toollery}
    BFCL-V4 non-live multiple results
  }
\centering
\resizebox{\textwidth}{!}{
\begin{tabular}{lcccccccc}
\hline \hline
\textbf{Model} &
  \textbf{$\leq$ 10} &
  \textbf{20} &
  \textbf{30} &
  \textbf{40} &
  \textbf{50} &
  \textbf{100} &
  \textbf{443}  \\ \hline \hline
\textbf{Qwen3-4B (Prompt)}                  & 95.50  & 99.50  & 97.50 & 96.50 & 95.50 & 90.00 & 24.25 \\
\textbf{Qwen3-4B (Prompt) + Toollery}       & \textbf{100.00} $\Uparrow$ & 97.50  & 98.00 $\Uparrow$ & 95.00 & 98.00 $\Uparrow$ & 91.50 $\Uparrow$ & 83.63 $\Uparrow$   \\ \hline
\textbf{Qwen3-8B (Prompt)}                  & 97.50  & 98.50  & \textbf{99.50} & 97.50 & 97.00 & 89.50 & 37.63  \\
\textbf{Qwen3-8B (Prompt) + Toollery}       & 99.50 $\Uparrow$  & \textbf{100.00} $\Uparrow$ & 97.50 & 96.00 & 98.00 $\Uparrow$ & 91.00 $\Uparrow$ & 80.63 $\Uparrow$    \\ \hline
\textbf{DeepSeek V3.2  (Prompt)}            & 86.00  & 99.50  & 97.00 & 94.50 & 96.50 & 85.50 & 70.25     \\
\textbf{DeepSeek V3.2  (Prompt) + Toollery} & 97.50 $\Uparrow$  & 97.50  & 95.00 & 98.00 $\Uparrow$ & \textbf{98.50} $\Uparrow$ & 89.50 $\Uparrow$ & 81.38 $\Uparrow$   \\ \hline
\textbf{A major proprietary LLM (Prompt)}               & 88.00  & 87.50  & 92.50 & 85.00 & 88.50 & 88.00 & 80.00          \\
  \textbf{A major proprietary LLM (Prompt) + Toollery}    & 88.00  & 89.50 $\Uparrow$  & 91.50 & 86.50 $\Uparrow$ & 95.00 $\Uparrow$ & 90.00 $\Uparrow$ & 82.00 $\Uparrow$   \\ \hline
  \textbf{A major proprietary LLM (FC)}                    & 79.50  & 69.50  & 65.50 & 72.00 & 70.50 & 63.50 & 58.63         \\
  \textbf{A major proprietary LLM (FC) + Toollery}         & 81.50 $\Uparrow$  & 77.00 $\Uparrow$  & 75.50 $\Uparrow$ & 73.00 $\Uparrow$ & 76.00 $\Uparrow$ & 69.50 $\Uparrow$ & 68.63 $\Uparrow$    \\ \hline
  
\textbf{Qwen3-8B (FC)}                      & 97.50  & 97.50  & 97.50 & \textbf{98.50} & 96.00 & 85.00 & 60.75   \\
\textbf{Qwen3-8B (FC) + Toollery}           & 99.50 $\Uparrow$  & 96.50  & 97.50 & 96.50 & 97.00 $\Uparrow$ & 90.50 $\Uparrow$ & 86.75 $\Uparrow$     \\ \hline
\textbf{Qwen3-235B (FC)}                    & 95.50  & 99.50  & 96.50 & 96.00 & 97.50 & 80.00 & 79.13             \\
\textbf{Qwen3-235B (FC) + Toollery}         & 96.00 $\Uparrow$  & 99.00  & 98.50 $\Uparrow$ & 98.00 $\Uparrow$ & 98.00 $\Uparrow$ & \textbf{93.50} $\Uparrow$ & \textbf{94.00} $\Uparrow$    \\ \hline \hline
\end{tabular}
}
\vskip -0.1in
\end{table*}

\section{Additional End-to-End Results}
\label{app:end-to-end-results}
Tables~\ref{bfcl_live_Result} and~\ref{bfclResult-toollery} provide additional BFCL-V4 end-to-end selection results beyond the main DeepSeek V3.2 comparison in Table~\ref{tab:atomic-tool-cost}. We vary the final selector, the decision interface, and the candidate-pool size before Toollery compression. The BFCL-V4 live multiple split contains 1,053 requests and 457 unique tools, while the non-live multiple split contains 200 requests and 443 unique tools; therefore, the final columns correspond to the complete candidate pools for the two splits.

These results test whether candidate compression remains useful beyond a single final LLM. Across prompt-based and function-calling settings, Toollery is most helpful when the candidate pool is large, where full-library prompting exposes the final LLM to more tool specifications and distractors. The trend supports the main result: Toollery improves the serving tradeoff by retrieving a compact candidate set before final LLM decision-making, rather than asking the final LLM to select from the full library.

\begin{table*}[t]
\centering
\small
\setlength{\tabcolsep}{2.5pt}
\renewcommand{\arraystretch}{1.08}
\begin{tabular}{
  >{\raggedright\arraybackslash}p{0.1\textwidth}
  >{\raggedright\arraybackslash}p{0.25\textwidth}
  >{\raggedright\arraybackslash}p{0.3\textwidth}
  >{\raggedright\arraybackslash}p{0.3\textwidth}
}
\hline \hline
\textbf{Dataset} & \textbf{Query / task} & \textbf{Gold capability} & \textbf{Top-5 retrieval} \\
\hline \hline
SkillRouter
& You need to fix build errors in a Java codebase.
The repository is located in ...
& \begin{tabular}[t]{@{}l@{}}
gt/maven-build-lifecycle\\
gt/maven-dependency-management\\
gt/maven-plugin-configuration
\end{tabular}
& \begin{tabular}[t]{@{}l@{}}
gt/analyze-ci\\
gt/uv-package-manager\\
development/ci-debug\\
other/creating-commit\\
other/github-pr-fix
\end{tabular} \\
\hline
SkillRouter
& Build an itinerary for the user according to the following requirements:
We require a 7-day travel itinerary for two leaving from Minneapolis ...
& \begin{tabular}[t]{@{}l@{}}
gt/search-accommodations\\
gt/search-attractions\\
gt/search-cities\\
gt/search-driving-distance\\
gt/search-flights\\
gt/search-restaurants
\end{tabular}
& \begin{tabular}[t]{@{}l@{}}
other/travel-planner-ai-claude-skills\\
other/tour-operations\\
gt/search-accommodations\\
development/travel-planner\\
other/travel-requirements-expert
\end{tabular} \\
\hline
SkillRouter
& In this task, you are given a clip of screen recording of a player playing the game Super Mario. Your goal is to analyze the key frames in this video and count how many coins / enemies / turtles ever showing up in each frame and write the results into a CSV file.
& \begin{tabular}[t]{@{}l@{}}
gt/ffmpeg\\
gt/image\_editing\\
gt/object\_counter
\end{tabular}
& \begin{tabular}[t]{@{}l@{}}
gt/object\_counter\\
gt/ffmpeg\\
gt/video\-frame\-extraction\\
other/video\-processing\\
data/video\-frame\-reader
\end{tabular} \\
\hline \hline
BFCL-live
& I'm having trouble understanding how to use the list\_files functionality. Help me.
& dartfx\_help
& \begin{tabular}[t]{@{}l@{}}
list\_files\\
list\_directory\_contents\\
help\\
list\_projects\\
adriel\_list\_projects
\end{tabular} \\
\hline
BFCL-live
& search flight prices from NYC to NEw Delhi use tools, finally respond in json
& search\_web\_tool
& \begin{tabular}[t]{@{}l@{}}
get\_trip\\
Flights\_4\_SearchRoundtripFlights\\
Flights\_4\_SearchOnewayFlight\\
render\_ui\_finish\\
Trains\_1\_GetTrainTickets
\end{tabular} \\
\hline
BFCL-live
& Find a recent news article about corona virus
& search\_web\_tool
& \begin{tabular}[t]{@{}l@{}}
search\_engine.query\\
get\_news\_report\\
HNA\_WQA.search\\
HNA\_WQA.query\\
HNA\_NEWS.search
\end{tabular} \\
\hline  \hline
\end{tabular}
\caption{Representative Toollery failure cases. Skill-routing failures often miss implementation-specific or complementary capabilities, while BFCL-live failures often arise from near-duplicate tool interfaces or underspecified requests.}
\label{tab:failure-cases}
\end{table*}

\section{Failure Case Analysis}
\label{app:failure-analysis}

Table~\ref{tab:failure-cases} shows representative retrieval failures from SkillRouter and BFCL-live. These cases suggest that many Toollery failures are near misses rather than unrelated retrieval errors. In SkillRouter, failures often reflect two intertwined issues: incomplete coverage of multi-capability workflows and mismatch between the benchmark's exact gold capability set and functionally similar retrieved capabilities. In BFCL-live, failures more often arise from fine-grained ambiguity among atomic tool interfaces, where the retrieved tools belong to the right tool family but do not match the exact gold interface.

\paragraph{Functionally similar capabilities can fail exact-gold matching.}
Several failures in Table~\ref{tab:failure-cases} are near misses under exact-label evaluation. In the SkillRouter build-repair case, the benchmark gold capabilities are Maven-specific, including \texttt{gt/maven-build-lifecycle}, \texttt{gt/maven-dependency-management}, and \texttt{gt/maven-plugin-configuration}. Toollery instead retrieves generic CI and debugging capabilities such as \texttt{gt/analyze-ci}, \texttt{gt/uv-package-manager}, and \texttt{development/ci-debug}, which are plausible for repairing build failures but do not match the exact gold labels. BFCL-live shows the same issue at the atomic-tool level: for a request asking for help with \texttt{list\_files}, Toollery retrieves \texttt{list\_files}, \texttt{list\_directory\_contents}, and generic help tools, but misses the annotated gold tool \texttt{dartfx\_help}. For flight-price and news-search requests, Toollery retrieves flight-specific, news-specific, or search-like tools, but misses the gold \texttt{search\_web\_tool}. These cases suggest that some retrieval failures reflect exact-interface or exact-label mismatch rather than completely unrelated retrieval.

\paragraph{Broad capabilities can substitute for decomposed gold labels.}
The travel-planning case illustrates a capability-granularity mismatch. The user request asks for an end-to-end itinerary, and Toollery retrieves broad travel-planning capabilities such as travel-planner-style agents, along with one exact gold capability, \texttt{gt/search-accommodations}. The benchmark gold set, however, decomposes the task into several fine-grained capabilities, including attraction search, driving-distance search, flight search, and restaurant search. Under exact recall, the broad travel-planner capability receives no credit for covering these subfunctions, even if it may be able to implement a complete itinerary workflow. A similar pattern appears in the Super Mario video-analysis case: Toollery retrieves broad video-processing or task-level capabilities together with \texttt{gt/object\_counter} and \texttt{gt/ffmpeg}, but misses the exact gold \texttt{gt/image\_editing}. These cases show that some failures reflect mismatch between the granularity of retrieved capabilities and the granularity of benchmark annotations.

Overall, the failure cases show two main boundary conditions for Toollery. First, in large capability libraries, multiple capabilities may be functionally overlapping, so exact-label evaluation can count feasible alternatives as failures. Second, for compositional tasks, a broad retrieved capability may cover an end-to-end workflow while the benchmark gold set decomposes the task into finer-grained capabilities. Future improvements should therefore consider interface-aware reranking for near-duplicate tools and staged retrieval for multi-capability workflows.

\end{document}